\documentclass[%
        compressed,
        ]{ieee}

\usepackage{graphicx}
\usepackage{subfigure}

\begin{document}

\title[Pigment Melanin: Pattern for Iris Recognition]{%
       Pigment Melanin: Pattern for Iris Recognition}

\author[HOSSEINI, ARAABI AND SOLTANIAN-ZADEH\'{A}R]
                        {Mahdi S. Hosseini\member{Student Member}, \authorinfo{M.\,S.\,Hosseini was with the School of Electrical and Computer Engineering, University of Tehran, Tehran, Iran, during this research. He is now with the Department of Electrical and Computer Engineering, University of Waterloo, Waterloo, ON, N2L 3G1. Phone: $+$1\,519\,888--4568 Ex.37459, e-mail: smhossei@uwaterloo.ca}%
                         Babak N. Araabi\member{Member},\authorinfo{B.\,N.\,Araabi is with the Control and Intelligent Processing Center of Excellence (CIPCE), School of Electrical and Computer Engineering, University of Tehran, North Kargar Ave., Tehran 14395-515, Iran. Phone: $+$98\,21\,8863--0024, e-mail: araabi@ut.ac.ir}%
\and{}and Hamid Soltanian-Zadeh\member{Senior Member} \authorinfo{H. Soltanian-Zadeh is with the Control and Intelligent Processing Center of Excellence (CIPCE), School of Electrical and Computer Engineering, University of Tehran, North Kargar Ave., Tehran 14395-515, Iran and the Image Analysis Lab., Radiology Dept., Henry Ford Hospital, Detroit, MI 48202, USA. Phone: $+$1\,313\,874--4482, e-mail: hamids@rad.hfh.edu}}

\journal{IEEE Trans.\ on Instrum.\ Meas.}
\titletext{, Vol.\ 59, No.\ 4, April\ 2010}
\loginfo{Manuscript received March 7, 2009.}


\maketitle               

\begin{abstract} 
Recognition of iris based on Visible Light (VL) imaging is a difficult problem because of the light reflection from the cornea. Nonetheless, pigment melanin provides a rich feature source in VL, unavailable in Near-Infrared (NIR) imaging. This is due to biological spectroscopy of eumelanin, a chemical not stimulated in NIR. In this case, a plausible solution to observe such patterns may be provided by an adaptive procedure using a variational technique on the image histogram. To describe the patterns, a shape analysis method is used to derive feature-code for each subject. An important question is how much the melanin patterns, extracted from VL, are independent of iris texture in NIR. With this question in mind, the present investigation proposes fusion of features extracted from NIR and VL to boost the recognition performance. We have collected our own database (UTIRIS) consisting of both NIR and VL images of 158 eyes of 79 individuals. This investigation demonstrates that the proposed algorithm is highly sensitive to the patterns of cromophores and improves the iris recognition rate. 
\end{abstract}

\begin{keywords}
Iris Biometrics, Visible-Light (VL), Near-Infrared (NIR), Pigment Melanin, Eumelanin, Shape Analysis, Image Enhancement, Regularized (Tikhonov) Filtering and Variational Binarization.
\end{keywords}

\section{Introduction}

\PARstart Iris recognition is one of the most reliable non-invasive methods of personal identification owing to the stability of the iris over one's lifetime. Pioneer work on iris recognition --as the basis of many commercial systems-- was carried out by Daugman~\cite{Daugman:93}. In this algorithm, 2D Gabor filters are adopted to extract oriented-based texture features corresponding to a given iris image. After Daugman, other researchers have contributed new methods to arrive at alternative algorithms with low computational burden, less \emph{SNR} and more compact codes, e.g. ~\cite{Wildes:97},~\cite{Boles:98},~\cite{Lim:2001},~\cite{Ma:2003},~\cite{Ma:2004} and~\cite{Ma2:2004}.

Most feature extraction methods have been implemented through multi-resolutional analysis, e.g. applying Laplacian pyramid construction with four different resolution levels~\cite{Wildes:97}; zero-crossing representation of 1D wavelet transform at various resolution levels of a virtual circle~\cite{Boles:98}; 2D wavelet decomposition ~\cite{Lim:2001}; and 1D Discrete Cosine Transform (1D-DCT)~\cite{Monro:2007} by zero crossing of adjacent patches, 1D-long and short Gabor filters ~\cite{Park:2007}, etc. Furthermore, alternative methods have been introduced based on the idea of local intensity variation in ~\cite{Ma:2004} and ~\cite{Ma2:2004}, entropy-based coding strategy in ~\cite{Proença:2007}, SVM-based learning approach in ~\cite{Vatsa:2008}, cross-phase spectrum in ~\cite{Kumar:2003}, two dimensional Fourier transform in ~\cite{Miyazawa:2008} and multi-lobe differential filters (MLDF) in ~\cite{Sun}. 

The majority of the benchmarks for iris recognition systems rely on Near-infrared (NIR) imaging rather than using visible-light (VL). This is due to the fact that fewer reflections from cornea in NIR imaging make the systems robust in recognition. However, compared to VL, NIR eliminates most of the related information in pigment melanin that scatters in the iris. This is due to the chromophore of the human iris, which has two distinct heterogeneous macromolecules called brown-black Eumelanin and yellow-reddish Pheomelanin~\cite{Liu:2005} and ~\cite{Meredith-Sarna:2006}.

Wielgus and Sarna~\cite{Wielgus-Sarna:2005} determined the amount of melanin in the human iris and the relative content of iron in the iridial melanin as a function of their color, shade, and the age of their donors by using electron spin resonance (ESR). This research proved that melanin in iris predominantly consisted of eumelanin with very similar chemical properties. Menon et. al.~\cite{Menon:92} discussed the amount of melanin in the iris pigment epithelium (IPE), which coats the posterior surface of the iris. This appears to contain mainly eumelnin with 97.6\% and 91.7\% in the case of blue-green and brown irides respectively~\cite{Prota:98}.

Eumelanin's radiative fluorescence under Ultra-Violet (UV) and VL excitation (e.g., $6\times 10^{-6}$ for VL)~\cite{Meredith:2006} influences the Charge-coupled device (CCD) sensor of the camera~\cite{ando-2006}. Studying the excitation-emission quantum yields of eumelanin, presented in Figure~\ref{Eumelanin}, shows that exciting this macromolecule under NIR firing leads to almost no emission of quantum yields where the related chromophors attenuate in NIR imaging.

However, processing of VL images is not as reliable as the NIR images due to the \emph{SNR} limitation and artifacts (e.g., reflection and shadows). This can weaken the procedure and, in order to overcome the limitations of VL imaging, new methods for robust feature extraction are suggested to help iris biometric systems to achieve accurate identifications. The demands for new developments to reach high accuracies in huge-size-databanks are not deniable. These demands are critical because there is additional information in the VL images. It is important to verify whether the features extracted from the NIR and VL can be combined in order to boost recognition rate as a fusion application instead of multi-biometric systems where the implementations are expensive.

It is noteworthy that in recent years working with the VL images has become more popular. Thornton, et al~\cite{Thornton:2007} introduced deformed Bayesian matching methodology and applied their algorithm to the CMU database (a VL database) and achieved reasonable results for both accuracy and computation time. Proenc¸a and Alexandre~\cite{Proenca-Alexandre:2007} divided the iris into six regions based on their own experience and applied Daugman's recognition method~\cite{Daugman:93} to compare the results.

In this paper, we introduce a different solution to the problem of VL imaging to explain melanin chromophore as a relevant pattern for shape analysis. The VL features should not be highly correlated with the NIR features to fuse the two modalities. Our preliminary results on the application of shape analysis to VL images was presented in ~\cite{Hosseini:2007}. In addition, our group at the University of Tehran (UTIRIS) published new results on this issue in ~\cite{Nima1:2008}, ~\cite{Nima2:2008} and ~\cite{Nima3:2009}.

The remainder of this paper is organized as follows. Section II briefly reviews the optical spectroscopy of the eumelanin in three main categories: Absorption, Emission, and Excitation. In Section 3, two categories, shape analysis and its invariant features and regularized Tikhonov filtering with logarithmic enhancement, are studied in order to increase the reliability of accessing to chromophore melanin in the iris. The proposed method for extraction of melanin patterns is introduced in Section 4 to produce robust shapes for the shape analysis techniques. Our experimental results for classification are evaluated in Section 5 using UBIRIS and CASIA databanks. Our data collection is then introduced where each iris has been captured in two different sessions, NIR and VL. We demonstrate that fusion of the extracted features in two sessions leads to higher classification accuracy. Finally, Section 6 presents discussions and concludes the work.

\section{Biological Roots: Eumelanin Optical Spectroscopy}
\begin{figure}
\centerline{
\subfigure[]{\includegraphics[width=.33\textwidth]{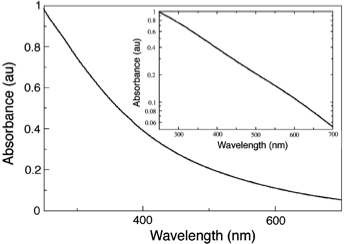}
\label{EumelaninAbs}}
\hfil
\subfigure[]{\includegraphics[width=.33\textwidth]{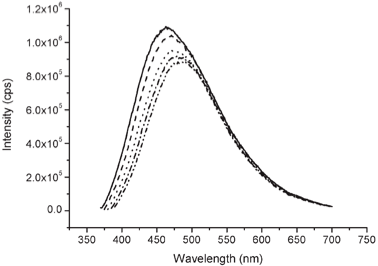}
\label{EumelaninFlo}}
\hfil
\subfigure[]{\includegraphics[width=.33\textwidth]{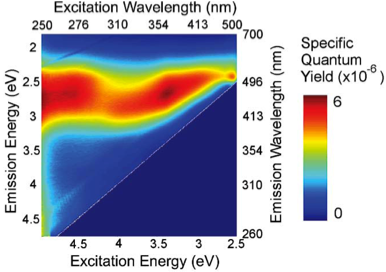}
\label{EumelaninQua}}}
\caption{(a) Eumelanin absorbance as a function of wavelength. The same data is shown in a semi-logarithmic plot in the insert demonstrating an excellent fit to the exponential model; (b) Eumelanin florescence emission under variety of excitation wavelengths from 360 nm (solid line) to 380 nm (inner dot-dashed line). Data are taken from; (c) Eumelanin specific quantum yield map: the fraction of photons absorbed at each excitation wavelength that are emitted at each emission wavelength. Data are taken from $\left[20\right]$.}
\label{Eumelanin}
\end{figure}

A brief review of the optical properties of eumelanin under different wavelength firings is presented in this section. Biophysists have introduced eumelanin as an intelligent predefined acid macromolecule~\cite{Pezzella:97}. Meredith et. al.~\cite{Meredith:2006} discussed the optical properties of the eumelanin in three main categories: absorbance, excitation, and emission. The eumelanin has the absorbance profile shown in Figure \ref{EumelaninAbs} absorbing lower wavelengths more than higher ones with maximum absorbance for ultraviolet wavelength. Its absorbance decreases exponentially and becomes almost relaxed in Near-Infrared ($750^{nm}$) with an extreme lower rate.

Regarding emission, the eumelanin emits light when stimulated by UV and VL; see Figure \ref{EumelaninFlo}. The emission depends on the excitation energy, in direct contrast with the Kasha's rule~\cite{Lakowicz:97} by disordering the mirror image rule for the organic chromophores, which states that the emission spectrum should approximately be a mirror image of the absorbance. The emission profile is close to the Gaussian function whose maximum is at about 460 nm (2.7 eV).

The excitation pattern at different wavelengths is another key to understanding the way eumelanin is stimulated. Recent studies~\cite{Nighswander:2005} have created a full ``radiative quantum yield map" showing the fate of each absorbed photon with respect to the emission; see Figure \ref{EumelaninQua}. This map shows the complexity and excitation energy dependency of the emission and the presence of a high wavelength bound on the emission. As it shown, the quantum yields for the NIR firing are attenuated while preserving quantities for the emission excited in VL. These emitted quantum yields, from 360 nm to 600 nm, can be captured by a CCD camera, and hence NIR imaging eliminates most pigment information in the iris where these chromophores mainly consist of the eumelanins.

\section{Shape Analysis and Image Enhancement}

The iris includes complex texture due to its pigments, blood vessels, crypts, contractile furrow, freckles, collarette, and pupillary frills~\cite{Westmoreland:99} which make it distinguishable from person to person. Extracting proper features which describe these patterns is useful for iris recognition. Current feature extraction methods from the iris are dominated by the wavelets and Gabor filters that were initially used in the first system introduced by Daugman [1] in 1993 (e.g.,~\cite{Wildes:97}-\cite{Monro:2007}). Several investigations evaluated the iris recognition methods using large databanks and enhanced the analysis performance by methods such as cascaded classifiers (e.g.,~\cite{Sun:2005}), Bayesian approaches (e.g.,~\cite{Schmid:2006}), and model-based methods (e.g.,~\cite{Zuo:2007}). 
The chromophore scattering makes the iris pattern complex and hard to explain. Thus the related feature to melanin cromophore should directly describe the related patterns. Such patterns can be presented in meaningful shapes to be analyzed by shape analysis techniques.

\subsection{Shape Analysis and Invariant Features}

Shape is a difficult concept to understand and is invariant through geometrical transformations such as translation, rotation, size changes, and reflection. A shape description can be achieved by functional explanation of a figure.  This is appropriate for many applications because of its advantages over other methods such as the following~\cite{Kindratenko}:

\begin{itemize}
        \item \textit{Effective data reduction}: a few coefficients of the approximating functions are frequently needed for a rather precise description.
        \item \textit{convenient description and intuitive characterization of complex forms}
\end{itemize}

Here, three distinct features are proposed and extracted from the iris images: \textit{Radius-Vector Function (RVF)}, \textit{Support Function (SF)} and \textit{Tangent-Angle Function (TAF)}.

\subsubsection{Radius-Vector Function (RVF)}

\begin{figure}
\centerline{\subfigure[]{\includegraphics[width=1.8in]{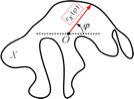}
\label{rv}}
\hfil
\subfigure[]{\includegraphics[width=1.8in]{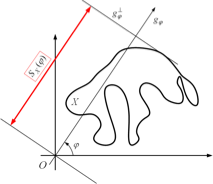}
\label{sup}}
\hfil
\subfigure[]{\includegraphics[width=1.8in]{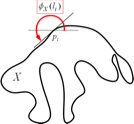}
\label{ta}}}
\centerline{\subfigure[]{\includegraphics[width=1.8in]{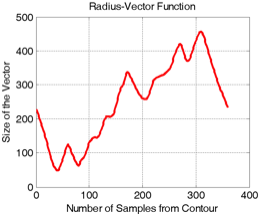}
\label{rvf}}
\hfil
\subfigure[]{\includegraphics[width=1.8in]{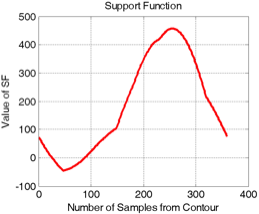}
\label{supf}}
\hfil
\subfigure[]{\includegraphics[width=1.8in]{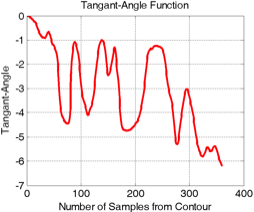}
\label{taf}}}
\centerline{\subfigure[]{\includegraphics[width=1.8in]{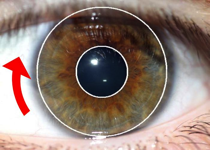}
\label{polar}}
\hfil
\subfigure[]{\includegraphics[width=1.8in]{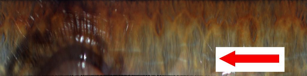}
\label{cartesian}}}
\caption{(a) Definition of radius-vector function; (b) Definition of Support Function; (c) Definition of Tangent Angle Function; (d) Sketch of Radius-Vector Function; (e) Sketch of Support Function; (f) Sketch of Tangent Angle Function; (g) Orientation of Iris in polar coordinates; (h) Translation of the mapped iris in the Cartesian coordinate system caused by rotation in the polar coordinate syatem.}
\label{ShapeAnalysis}
\end{figure}

A reference point $O$ in the interior of the figure $X$ is selected. Next, an appropriate reference line $l$ crossing the reference point $O$ is chosen parallel to the x-axis. The radius-vector function $r_X(\varphi)$ is defined by the distance between the reference point $O$ and the crossing point with the contour in the direction of $\varphi$ ($0 \leq \varphi \leq 2 \pi$) scanning the contour in anticlockwise direction, see Figure \ref{rv} and \ref{rvf}. The contour is obtained by labeling separate contours in an image and calling the related graph. This can be done in MATLAB using `\texttt{bwlabel}' function. The contour is been scanned from an arbitrary point to cycle the closed contour in anti-clockwise direction.

We use all points of the figure as potential features and normalize the radius-vector function. For example, in Fig. 2d, the number of points is normalized to 360 points. The RVF $r_X(\varphi)$ has the following features:

\begin{itemize}
        \item Invariant under translation: $r_{X+t}(\varphi)=r_X(\varphi)$ where $X+t$ is $X$ translated by a vector $t$.
        \item Depends on figure $X$'s size: $r_{\lambda X}(\varphi)=\lambda r_{X}(\varphi))$ where $\lambda X$ is the figure $X$ zoomed by a factor $\lambda$. Nevertheless, this limitation can be avoided by normalizing the radius function.
        \item Variant under reflection: flipping the image through the vertical axis (image in mirror) which does not happen in iris imaging.
        \item Depends on the orientation of the figure $X: r_Y(\varphi)=r_{X}(\varphi-\alpha)$ where $Y$ is figure $X$ rotated by an angle $\alpha$
\end{itemize}

The last two properties will not influence iris characterization since rotation in the polar coordinate system equals translation in the Cartesian coordinate system; see Figure \ref{polar} and \ref{cartesian}. This allows us to start to scan the contour from an arbitrary point $p_0$.

\subsubsection{Support Function (SF)}

For a figure $X$, let $g_{\varphi}$ be an oriented line through the origin $O$ with direction $\varphi$ ($0\leq \varphi \leq 2 \pi$). Let $g^{\varphi}_{\bot}$ be the line orthogonal to $g_{\varphi}$ so that the figure $X$ lies completely in the half-plane determined by $g^{\varphi}_{\bot}$ with $g^{\varphi}_{\bot} \cap X \neq O$, which is opposite to the direction of $g_{\varphi}$. The absolute value of the support function equals the distance from $O$ to $g^{\varphi}_{\bot}$. The support function (SF) $S_{X}(\varphi)$ is negative if the figure lies behind $g^{\varphi}_{\bot}$ as seen from the origin. If $O$ belongs to the figure $X$, then $S_{X}(\varphi)\geq 0$ for all $\varphi$, see figure \ref{sup} and \ref{supf}. All of the four features listed above for the RVF are applicable to the SF. The SF can be calculated by the following equation~\cite{Kindratenko}:
\begin{equation}
S_{X}(\varphi)=\max_{0\leq l\leq L}\left[x_{X}(l)cos(\varphi)+y_{X}(l)sin(\varphi)\right]
\end{equation}

where $L$ is the perimeter of the figure $X$ and each point $x_{X}(l),y_{X}(l)$ of the contour of $X$ is associated with a number $l$. The relation between the polar and Cartesian coordinates of the figure can be written as:
\begin{equation}
\begin{array}{l}
x_{X}(l)=r_{X}(l)cos(\varphi_l) \\
y_{X}(l)=r_{X}(l)sin(\varphi_l)
\end{array}
\end{equation}

The SF is robust to the local distortions of the boundary because these distortions are small relative to the object's perimeter.

\subsubsection{Tangent-Angle Function (TAF)}
The third proposed feature is the angle of the tangent line at each boundary point. Let us assume that the perimeter of the figure $X$ is $L$. Every point $p_l$ on the boundary (contour) of $X$ can be associated by a number $l$ ($0\leq l\leq L$). The initial point $p_0$ is considered as the starting point, and the boundary points are considered in the anticlockwise direction. The orientation of the tangent line at the point $p_l$ is denoted by $\phi_{X}(l)$ and called tangent-angle function; see Figure \ref{ta} and \ref{taf} ~\cite{Stoyan:95}.

\subsection{Image Enhancement}

Iris images captured in Visible-Light (VL) are affected by noise. Most of the iris areas can be enhanced and observed correctly by appropriate enhancement tools. After iris extraction, the image  is enhanced by the homomorphic filtering using a logarithmic transformation and a Tikhonov filter as explained next.

The image intensities consist of two components: a) the source illumination incident on the scene being viewed and b) the reflectance of the objects in the scene~\cite{Gonzalez:2002}. These components are denoted by $i(x,y)$ and $r(x,y)$, respectively.  The multiplication of these two components is the image intensity $f(x,y)$, i.e.,
\begin{equation}
f(x,y)=i(x,y)r(x,y)
\end{equation}
where $0\leq i(x,y)\leq \infty$ and $0\leq r(x,y)\leq 1$. The two components can be separated by taking the logarithm of $f(x,y)$:
\begin{equation}
\begin{array}{l}
\log\left(f(x,y)\right)=\underbrace{\log\left(i(x,y)\right)}_{I}+\underbrace{\log\left(r(x,y)\right)}_{R}\\
f^{Enhanced}=\exp\left[\textbf{Normalized}\left(I+R\right)\right]
\end{array}
\end{equation}

\begin{figure}
\centerline{\subfigure[]{\includegraphics[width=1.5in]{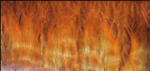}
\label{ColorIris}}
\hfil
\subfigure[]{\includegraphics[width=1.5in]{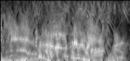}
\label{GreyIris}}
\hfil
\subfigure[]{\includegraphics[width=1.5in]{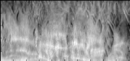}
\label{LogEnhance}}
\hfil
\subfigure[]{\includegraphics[width=1.5in]{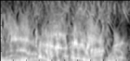}
\label{TikhEnhance}}}
\centerline{\subfigure[]{\includegraphics[width=2in]{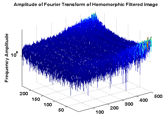}
\label{fft}}
\hfil
\subfigure[]{\includegraphics[width=2in]{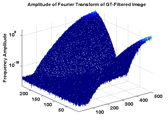}
\label{fftEnhance}}
\hfil
\subfigure[]{\includegraphics[width=2in]{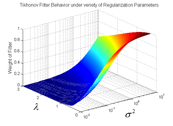}
\label{TikhonovFilt}}}
\caption{(a) Original color image of the iris; (b) Gray scale image; (c) Enhanced image by homomorphic filtering; (d) Filtered image through frequency using regularized Tikhonov value and visual improvement; (e) Amplitude of the Fourier transform of the image in c; (f) Amplitude of the Fourier transform of the Tikhonov filtered image in d; (g) Tikhonov filter for a variety of regularization parameters.}
\label{Enhancement}
\end{figure}

The $R$ element of an iris image is highly correlated with the back flash of the light from the cornea.  By normalizing the logarithm of the image intensities simply scaling $\left(I+R\right)\in[0,1]$ and taking the exponential of the normalized value, the reflectance variations can be attenuated; see Figure \ref{ColorIris}--\ref{LogEnhance}.

The VL iris images consist of light reflection and shadows. The homomorphic filtering enhances the specularities in the iris. However, there is still high frequency noise that can be miss-classified as chromophors. The nature of high frequency is a sparse representation in the Fourier domain, and proper regularization filtering can delete such noises. The Regularaized Tikhonov Filter~\cite{Horn-Holfort} is a powerful tool to separate such smooth chromophore variations from high frequency noise. Equation \ref{TikhonovEq} is known as general Tikhonov regularization:
\begin{equation}
f_{\lambda}=\arg\min_{f}\left\{{\left\|\textbf{A}f-b\right\|}^{2}_{2}+\lambda^{q}{\left\|\textbf{L}f\right\|}^{q}_{q}\right\}
\label{TikhonovEq}
\end{equation}

where $q$ is the space norm (here $q=2$), $\textbf{L}$ is a linear operator on the original signal. So the added term is called the regularization term and The problem is to reconstruct the sharp unknown image $\textbf{F}$ from a given blurred image $\textbf{B}$ ($f$ and $b$ are the vectorized tensors, $vec(\textbf{F})$). The added term reduces the effect of noise on the results. The regularization parameter $\lambda$ (here $\lambda=0.8$) controls the \textit{a priori} knowledge regarding the residual norm. The final solution can be expressed as:
\begin{equation}
F_i=\frac{\sigma^{2}_{i}}{\sigma^{2}_{i}+\lambda^2}
\end{equation}
where $\sigma_i$ is a singular value of the blurring matrix \textbf{A}. Figure \ref{TikhonovFilt} shows the behavior of the Tikhonov filter for a range of regularization parameters ($0\leq \lambda\leq 3$). The practical implementation of the above filter is more convenient in Fourier domain which is introduced by Hansen et al~\cite{Hansen:2006}. In this method, the singular values are considered by the elements of the Fourier transform of the image and a point spread function as a smoothing operator, e.g., the Gaussian function. Then, the Tikhonov regularization is done by:
\begin{equation}
\textbf{B}^{GT}=
\mathcal{F}^{-1}_{2D}\left\{
\frac{\mathcal{F}_{2D}\left\{\textbf{PSF}\right\}^{\ast} \cdot \mathcal{F}_{2D}\left\{\textbf{B}\right\}}{\mathcal{F}_{2D}\left\{\textbf{PSF}\right\}^{\ast} \cdot \mathcal{F}_{2D}\left\{\textbf{PSF}\right\}+\lambda^2}
\right\}
\end{equation}

where $\mathcal{F}_{2D}$ stands for the 2-Dimensional Fourier transform operator, $\textbf{B}$ the input image, $\textbf{PSF}$ point spread function operator (variance of the operator here is $\sigma^2=25$) and $\textbf{B}^{GT}$ is the output of the generalized-Tikhonov filter.

\section{Proposed Method}
In our proposed method we rely on pattern of melanin pigments and the shapes generated by them in VL iris images, as individual identifiers. In this section we define and explain how to extract our proposed features.

\subsection{Image Binarization}
\begin{figure}
\centerline{\includegraphics[width=5.5in]{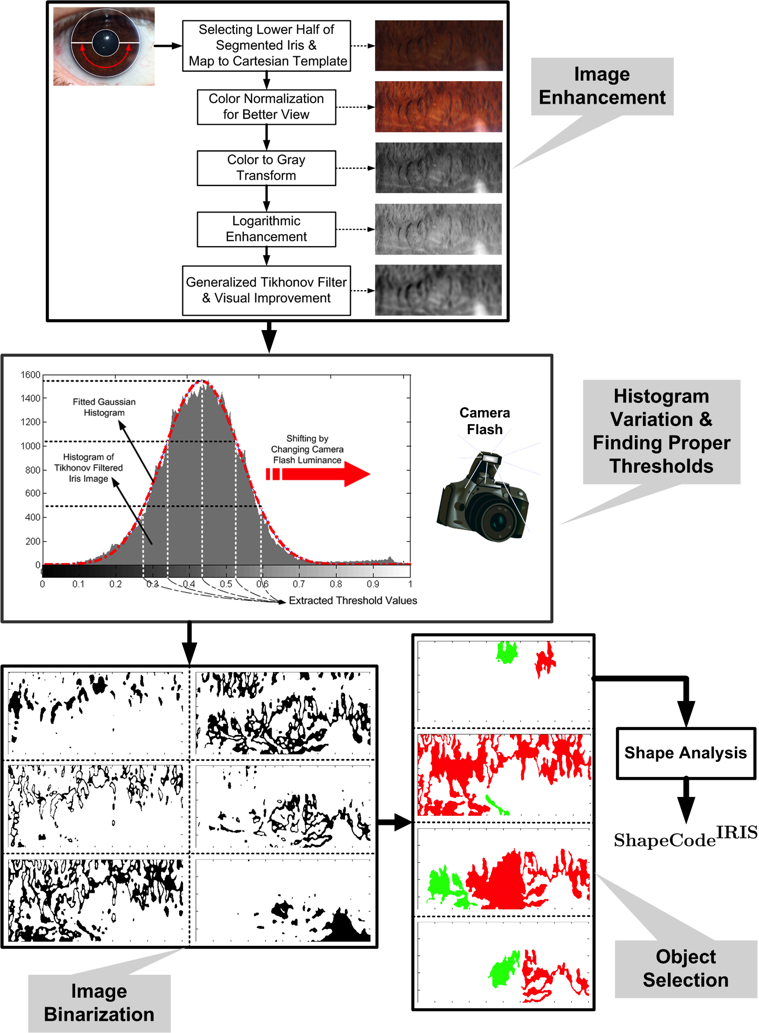}}
\caption{At first, captured iris image is enhanced by logarithmic and regularized Tikhonov filter. Next, A Gaussian distribution is fitted to the histogram of image. The filtered image is binarized after deriving adapted threshold values using histogram of grey intensities. Here, five threshold values calculated along intersection of horizontal line of intensity value with fitted Gaussian distribution. Then the filtered image is binarized based on derived threshold values and massive objects/patches are selected for shape analysis.}
\label{BlockDiagram}
\end{figure}

A block diagram of the proposed algorithm for feature extraction is shown in Figure \ref{BlockDiagram}. A simple approach to produce binarized-shapes is to cut the surface of an image by fixed threshold values. However, reflections and specularities from the cornea affect the image and make the platform sensitive to light intensities. The amount of luminance of camera-flash directly integrates with grey values and can shift the histogram of image. This affects the binarization of image and changes the pattern of generated shapes. Nonetheless, this amount of illumination is an injected-value and we can adaptively threshold the intensity surface of image by sliding along the injected-luminance to achieve robust threshold values, see Figure \ref{BlockDiagram} for \emph{Histogram Variation \& Finding Proper Threshold} block.

To generate robust shapes, a bell-shaped Gaussian function is fitted to the histogram. The histogram is considered Gaussian because of the limited variation of color in a unique iris. As an exception in some images, the color of the iris consists of different shadows which cause more than one peak in the histogram. Nevertheless, among the defined peaks, only one is dominant and all of them can be fitted properly with a Gaussian, see Figure \ref{BlockDiagram} for \emph{Histogram Variation \& Finding Proper Threshold} block. The tip of Gaussian defines one of the thresholds. Two horizontal lines that divide the Gaussian from zero level to tip into three equal levels, provide four more thresholds, Figure \ref{BlockDiagram}. Equation \ref{Binarization} defines the binarization:
\begin{equation}
\textbf{SlicedImage}^{i}=
\left\{
\begin{array}{l}
\textbf{I}\left(t^{i-1}\leq \textbf{I}\leq t^{i}\right)=1\\
\textbf{I}\left(\textbf{I}<t^{i-1}~~\&~~\textbf{I}> t^{i}\right)=0
\end{array}
\right.
\label{Binarization}
\end{equation}

where $t^i$ relates to $i^{th}$ extracted threshold value, $t^0=0$, $t^6=1$, and $i=1, 2, \ldots, 6$. In fact, the number of gray values in intensity histogram between two consecutive thresholds represents the shape formed by intensities lying between those two thresholds. One may conjecture that these shapes are a reflection of Eumelanin chromophores patterns discussed in Section II. It is possible to define more threshold values, similar to those defined, particularly around the mean of Gaussian distribution. However, one should bear in mind that this will increase the size of the code strip. The other possibility is to concentrate around the mean of Gaussian with a fixed number of threshold values, which can be problematic when the Gaussian function is rather flat with a large variance, since we might lose other areas that are not small in quantity.

\subsection{Object Selection and Matching Algorithm}

The binarized images (see figure \ref{BlockDiagram} for \emph{Image Binarization} block) contain several disconnected objects. Each shape represents the related intensity values between two finite thresholds. The bigger objects resukt in more robust extracted features. This is because of confliction between noises and true Eumelaning shades. To mitigate this conflict, we select larger shapes in order to increase the probability of related pigments in represented shapes. Two disconnected objects in each binarized template are considered. The first and the last templates are eliminated (see Figure \ref{BlockDiagram} for \emph{Object Selection} block) because they show the lowest and the highest intensities (often relate to shadows and reflected noises).

Three features, defined in Section 3: RVF, SF and TAF, are applied to the selected objects. In sum, four templates and two objects per template is considered. With three features per object, we have 24 code-strips, where altogether represent every iris image. Figure \ref{ShapeCode} shows how the features are sorted and inserted together.

\begin{figure}
\centering
\begin{math}
\textbf{RVF}=\left[\begin{array}{l}
\textbf{RVF}^{1}_{1}\\
\textbf{RVF}^{2}_{1}\\
\vdots\\
\textbf{RVF}^{1}_{4}\\
\textbf{RVF}^{2}_{4}
\end{array}
\right],
\textbf{SF}=\left[\begin{array}{l}
\textbf{SF}^{1}_{1}\\
\textbf{SF}^{2}_{1}\\
\vdots\\
\textbf{SF}^{1}_{4}\\
\textbf{SF}^{2}_{4}
\end{array}
\right],
\textbf{TAF}=\left[\begin{array}{l}
\textbf{TAF}^{1}_{1}\\
\textbf{TAF}^{2}_{1}\\
\vdots\\
\textbf{TAF}^{1}_{4}\\
\textbf{TAF}^{2}_{4}
\end{array}
\right]
\Rightarrow
\begin{array}{c}
\textbf{ShapeCode}^{\textbf{Iris}}= \\
\left[\begin{array}{l}
\textbf{RVF}\\
\textbf{SF}\\
\textbf{TAF}
\end{array}
\right]_{\textbf{24}\times \textbf{N}}
\end{array}
\end{math}
\caption{Definition of sorting the produced features, e.g. $\textbf{SF}^{i}_{j}$ defines the $i^{th}$ object from $j^{th}$ template. $\textbf{N}$ is the number of sample per feature.}
\label{ShapeCode}
\end{figure}

Unlike the other generated binarized code for iris, e.g. $\texttt{IrisCode}\copyright$~\cite{Daugman:93}, this feature code is defined as a gray value. The number of samples per feature is an important challenge which can influence the accuracy of matching algorithm. The size of iris image which is mapped to Cartesian coordinates has $256\times512$ pixels from captured $600\times800$ pixels iris image (we select the lower half ring of iris). We found that $\textbf{N}=100$ samples are sufficient to describe each feature. Each gray value of $\textbf{ShapeCode}^{\textbf{Iris}}$ is defined by 8-bit memory ($2^8=256$ level) ), so the size of each code for an iris image is calculated by:
\begin{equation}
Size~of\left\{\textbf{ShapeCode}^{\textbf{Iris}}\right\}=\textbf{M}\times \textbf{N}\times \textbf{B}
\end{equation}

where $\textbf{M}$ is the number of features (here equals 24), $\textbf{N}$ the number of samples used to introduce each feature as discrete signal (here equals 100) and $\textbf{B}$ is the number of bits to define each gray value (here equals 8). To compare two shape codes, the nearest-neighbor method from~\cite{Monro:2007} is used to calculate product of the sums (POS) of individual sub-feature Hamming distances ($\textbf{HD}$). In addition, the third parameter is added to define the number of bits regarding the depth of levels:
\begin{equation}
\textbf{HD}=\left(\prod^{\textbf{M}}_{i=1} \frac{\sum^{\textbf{B}}_{k=1}\sum^{\textbf{N}}_{j=1} \left(SF^{1}_{ijk}\oplus SF^{2}_{ijk}\right)}{\textbf{N}\times \textbf{B}}\right)^{1/\textbf{M}}
\end{equation}

where $SF$ stands for $SubFeature$ The equation defines the number of code-strips. In the related equation, the two sub-features are XORed and normalized by the size of code. So the Hamming distance (HD) is normalized between [0, 1]. We used POS-HD to make our proposed method comparable with other iris recognition methods, which mostly used this method for matching.

Proposed method can be challenged by occlusion of pupil or partial iris information. In this case, the conversion of extracted iris in Cartesian coordinate using fixed mask size ($256\times 512$) can blurs the image where information in radial axis is re-generated. This effect can influence the accuracy of the classification by generating distorted shapes. Another difficulty may be caused by fake ID, e.g. contact lenses. In the case of optical contact lenses, proposed method is resistant to such distortions, which is merely an additional layer to Cornea.

\section{Experimental Reuslts}
In all experiments, Daugman's method~\cite{Daugman:93} have been used to extract irises and convert them from polar coordinate to Cartesian. We used threshold pre-processing prior to extraction to maximize the contrast between iris and non-iris regions. This threshold is selected adaptively using gray image histogram. We considered the lower half of iris to reduce the occlusion effects. Such irregularity will damage the shape of extracted contour and can affect curves driven from feature code. However, we can consider the whole iris ring and increase the reliability of the system.
\subsection{Study on UBRIS and CASIA V.1}
The proposed algorithm is evaluated on two famous datasets in this section. The first dataset is UBIRIS~\cite{UBIRIS}, consists of 1877 images of 241 individuals, 1214 images from the first and 663 images from the second session. Each person has five captured images in different time sequences. This dataset is highly noisy due to several noise factors such as eyelids, eyelashes, glasses, the pupil (e.g. distortions through iris segmentation), motion blur, lighting and reflections~\cite{Proenca-Alexandre:2007}. The second dataset used is CASIA V.1~\cite{CASIA} which contains 756 NIR images from 108 eyes pertaining to 80 individuals. Each eye holds seven captured images presented in two different sessions with different time sequences (one month interval).

\begin{table}
\renewcommand{\arraystretch}{1}
\caption{1-FAR Classification Results for Different Scenarios of Train (Tr) and Test (Te) for UBIRIS \& CASIA}
\centering
\begin{tabular}{c|c|c|c|c|c}
 \hline
 \hline
  & \multicolumn{3}{c}{UBIRIS} & \multicolumn{2}{c}{CASIA} \\
 \hline
 & Scen. & S.\#1 & S.\#2 & Scen. & Vr. \#1 \\
 \hline
 1 & 4Tr, 1Te & 95.90\% & 94.92\% & 6Tr, 1Te & 65.74\% \\
 \hline
 2 & 3Tr, 2Te & 94.02\% & 93.70\% & 5Tr, 2Te & 60.65\% \\
 \hline
 3 & 2Tr, 3Te & 90.50\% & 89.42\% & 4Tr, 3Te & 52.47\% \\
 \hline
 4 & 1Tr, 4Te & 81.70\% & 81.37\% & 3Tr, 4Te & 47.69\% \\
 \hline
 5 & - & - & - & 2Tr, 5Te & 43.15\% \\
 \hline
 6 & - & - & - & 1Tr, 6Te & 30.25\% \\
 \hline
 \hline
\end{tabular}
\label{UbirisCasia}
\end{table}
Regarding UBIRIS and CASIA V.1, four and six different Scenarios could be introduced for the classification problem illustrated in Table \ref{UbirisCasia}. The ``\emph{train}'' and ``\emph{test}'' data are selected randomly. The related scenarios come from considering more than one image for training purposes. If we consider 'k' images ($1\leq k\leq$n) out of '\emph{n}' images per individuals (here $n=5$ for UBIRIS) for training purpose, then we will have '\emph{nk}' choices for testing purpose. The pattern of such selection is uniformly at random and there is no priority in train-image selection. Having more than one image for training purpose, we compare the test image to all train ones and chose the nearest neighbor in that class.

Melanin chromophore in an NIR image attenuates while it preserves in VL. These chromophores scatter in iris in a gradual variation manner. The proposed feature extraction method uses a variation approach to binarize iris image. It fits Gaussian profile on image histogram and extracts the related threshold values for binarization. Indeed, the noise factors in VL images are distinguishable from Melanin chromophore scatter, since such noises are represented in high frequencies by nature, while such chromophores distribute in lower frequencies band. Tikhonov regularization can solve the problem by ignoring high frequencies. As for NIR images such as CASIA.v1, due to disappearance of such chromophores, the information remained in such images are about textures, where usually contain high frequency information. Tikhonov filter deletes much of high frequency information; and since there is no chromophore related information, there remains very low information to be analyzed. That is why our method is not suitable for NIR images but it is a strong tool to code VL images.

\begin{figure}
\centerline{
\subfigure[]{\includegraphics[width=2.5in]{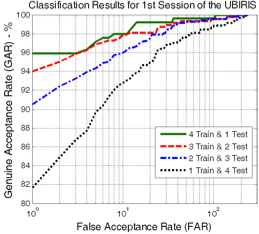}
\label{farUBIRIS1}}
\hfil
\subfigure[]{\includegraphics[width=2.5in]{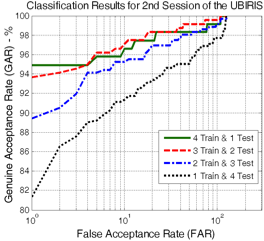}
\label{farUBIRIS2}}}
\centerline{
\subfigure[]{\includegraphics[width=2.5in]{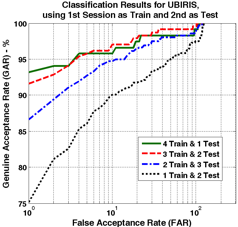}
\label{farUBIRIS}}
\hfil
\subfigure[]{\includegraphics[width=2.5in]{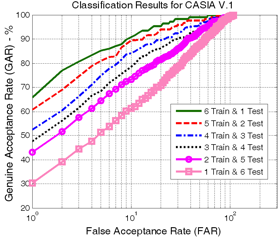}
\label{farCASIA}}}
\centerline{
\subfigure[]{\includegraphics[width=2in]{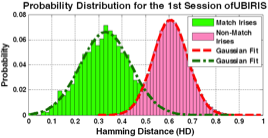}
\label{pdfUBIRIS1}}
\hfil
\subfigure[]{\includegraphics[width=2in]{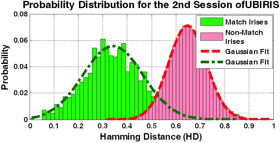}
\label{pdfUBIRIS2}}}
\centerline{
\subfigure[]{\includegraphics[width=2in]{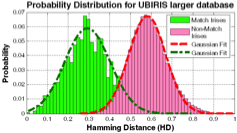}
\label{pdfUBIRIS}}
\hfil
\subfigure[]{\includegraphics[width=2in]{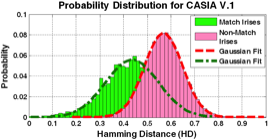}
\label{pdfCASIA}}}
\caption{(a) Classification on 1st Session of UBIRIS; (b) Classification on 2nd Session of UBIRIS; (c) Classification results on larger database of UBIRIS(1st + 2nd); (d)Classification results on CASIA V.1; (e) Probability distribution UBIRIS.S1; (f) Probability distribution on UBIRIS.S2; (g) Probability distribution on UBIRIS.(S1+S2); (h) Probability distribution on CASIA V.1}
\label{UBIRISCASIA}
\end{figure}

The identification accuracy of the system is illustrated by the False Acceptance Rate (FAR) in figure \ref{UBIRISCASIA} for both datasets. Comparing classification results in both sessions of UBIRIS (figure \ref{farUBIRIS1} and \ref{farUBIRIS2}) and mixture of both sessions, one can see that the efficiency of the proposed algorithm is consistently high through noise factors, where the second session of UBIRIS contains more noise factors than the first session, while the same accuracy is achieved in both sessions. The number of subjects in the first session is almost 2 times bigger than the second one, that is 241 and 125 individuals respectively (the number of subjects in 2nd Session of UBIRIS is 132, but we could only extract iris accurately from 125 subjects). Proenc¸a and Alexandre~\cite{Proenca-Alexandre:2007} applied Daugman's method~\cite{Daugman:93} on UBIRIS dataset, where they selected only 80 .from 241 individuals to maintain the size of the dataset comparable with that of CASIA. They did not, however, explain how they selected these individuals. Probability distribution of intraclass and interclass hamming distances for all 241 individuals are shown in Figure \ref{pdfUBIRIS1}, \ref{pdfUBIRIS2} and \ref{pdfUBIRIS}. 

\subsection{UTIRIS Database Collection}

The results of UBIRIS led to good accuracy due to the knowledge that the following database is a hard set to implement recognition algorithms. In contrast, the results for CASIA are poor, given the fact that CASIA contains fewer noisy images compared to UBIRIS. Shape analysis algorithm is highly dependent on pigment melanin. In order to study the effect of the above phenomenon, we decided to gather our own database in the University of Tehran, called UTIRIS, which consists of VL and NIR Iris images taken from the same individuals. UTIRIS consists of two sessions with 1540 images, half of which captured in a visible-light session and the other half captured in a NIR illumination session. Both sessions hold 158 eyes pertaining to 79 individuals (Right and Left eyes), Table \ref{UTIRISdatabase} explains the related dataset.

\begin{table}
\renewcommand{\arraystretch}{1}
\caption{UTIRIS Dataset Definition}
\centering
\begin{tabular}{ccccc}
 \hline
 \hline
 & \multicolumn{2}{c}{VL-Session} & \multicolumn{2}{c}{NIR-Session} \\
 \hline
 \hline
 Camera & \multicolumn{2}{c}{CANON EOS 10D + MACRO LENS} & \multicolumn{2}{c}{ISG LIGHTWISE LW} \\
 \hline
 & Right Eye & Left Eye & Right Eye & Left Eye \\
 \hline
 {Number of Individuals} & 79 & 79 & 79 & 79 \\
 \hline
 {Number of Images per Iris} & 5$\pm$1 & 5$\pm$1 & 5$\pm$1 & 5$\pm$1 \\
 \hline
 {Total number of images} & \multicolumn{2}{c}{770} & \multicolumn{2}{c}{770} \\
 \hline
 \hline
\end{tabular}
\label{UTIRISdatabase}
\end{table}

\begin{figure}
\centerline{
\subfigure[]{\includegraphics[width=1.5in]{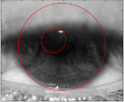}
\label{noise1}}
\hfil
\subfigure[]{\includegraphics[width=1.5in]{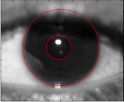}
\label{noise2}}
\hfil
\subfigure[]{\includegraphics[width=1.5in]{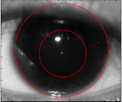}
\label{noise3}}
\hfil
\subfigure[]{\includegraphics[width=1.5in]{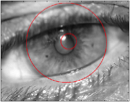}
\label{noise4}}}
\caption{UTIRIS noisy sample images containing interclass variations of focus, reflection, eyelashes, eyelids and shadows including Daugman's iris segmentation results on each}
\label{noiseUTIRIS}
\end{figure}

Images in VL Session have been captured in high resolution with 3 megapixels where they have been down-sampled by a factor of 2 in each dimension to have the same size as NIR captured images. As mentioned in the beginning, Daugman segmentation method is used to extract the iris. The iris radius is approximately extracted about 120 pixels in polar coordinates, where all mapped to 150 pixels in Cartesian rectangular pixels with $1^\circ$ circular arc to resolve in Cartesian coordinates. The rectangular region is ($150 \times 300$) for lower-half iris. Although the irises have been captured in high resolution, the images are highly noisy due to focus, reflection, eyelashes, eyelids and shadow variations, all of which make the iris matching a more difficult task. Figure \ref{noiseUTIRIS} shows some highly noisy difficult to code -as a result difficult to match- sample images along with segmentation results.

\begin{figure}
\centerline{
\subfigure[VL-1]{\includegraphics[width=1.5in]{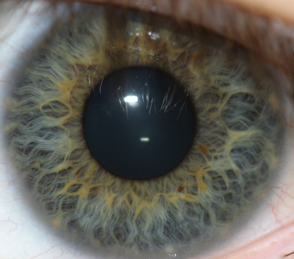}
\label{vl1}}
\hfil
\subfigure[NIR-1]{\includegraphics[width=1.5in]{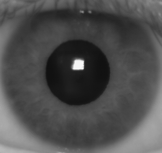}
\label{inf1}}
\hfil
\subfigure[VL-2]{\includegraphics[width=1.5in]{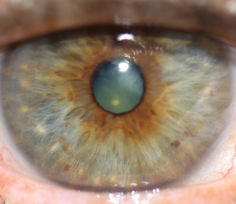}
\label{vl2}}
\hfil
\subfigure[NIR-2]{\includegraphics[width=1.5in]{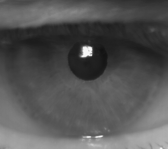}
\label{inf2}}}
\centerline{
\subfigure[VL-3]{\includegraphics[width=1.5in]{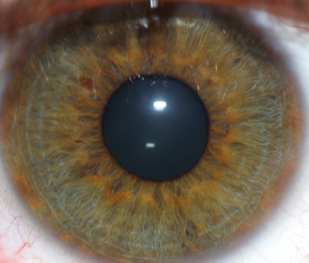}
\label{vl3}}
\hfil
\subfigure[NIR-3]{\includegraphics[width=1.5in]{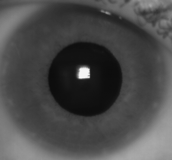}
\label{inf3}}
\hfil
\subfigure[VL-4]{\includegraphics[width=1.5in]{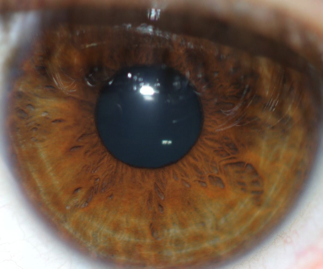}
\label{vl4}}
\hfil
\subfigure[NIR-4]{\includegraphics[width=1.5in]{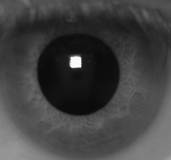}
\label{inf4}}}
\centerline{
\subfigure[VL-5]{\includegraphics[width=1.5in]{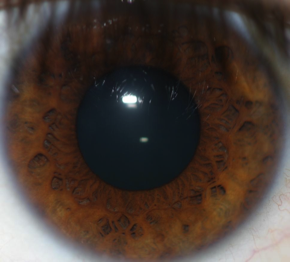}
\label{vl5}}
\hfil
\subfigure[NIR-5]{\includegraphics[width=1.5in]{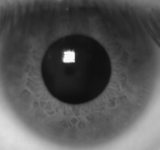}
\label{inf5}}
\hfil
\subfigure[VL-6]{\includegraphics[width=1.5in]{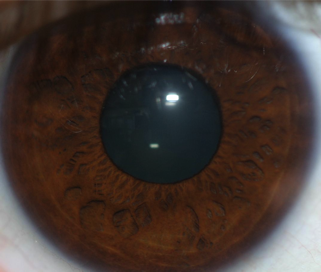}
\label{vl6}}
\hfil
\subfigure[NIR-6]{\includegraphics[width=1.5in]{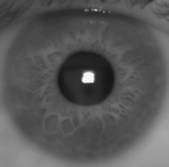}
\label{inf6}}}
\caption{Six different iris image from UTIRIS. from green to red and their appearance under both VL and NIR stimulation }
\label{UTIRISCOLOR}
\end{figure}

As mentioned, iris pigment epithelium (IPE) contain mainly eumelnin with 97.6\% and 91.7\% in the case of blue-green and brown irises respectively~\cite{Prota:98}. As it is shown in Figure \ref{UTIRISCOLOR} among six different varieties of iris colors, irises with green-blue colors lose more information in NIR imaging than brown, where the reason goes back to the higher content of eumelanin.

\begin{figure}
\centerline{
\subfigure[]{\includegraphics[width=2.5in]{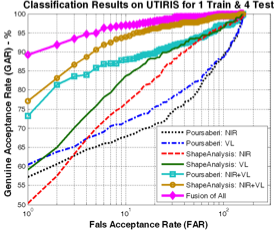}
\label{utiris1}}
\hfil
\subfigure[]{\includegraphics[width=2.5in]{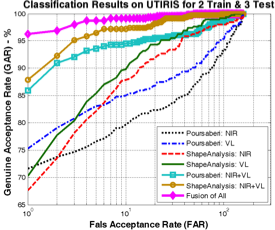}
\label{utiris2}}}
\centerline{
\subfigure[]{\includegraphics[width=2.5in]{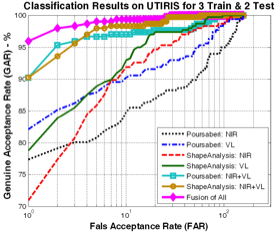}
\label{utiris3}}
\hfil
\subfigure[]{\includegraphics[width=2.5in]{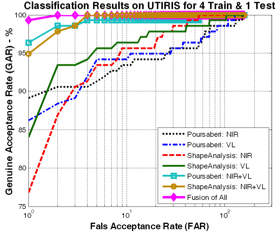}
\label{utiris4}}}
\centerline{
\subfigure[]{\includegraphics[width=2in]{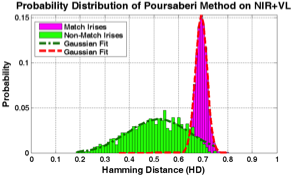}
\label{pdfAhmad}}
\hfil
\subfigure[]{\includegraphics[width=2in]{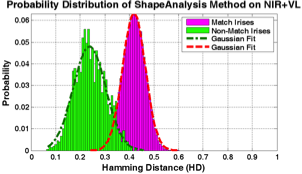}
\label{pdfShape}}
\hfil
\subfigure[]{\includegraphics[width=2in]{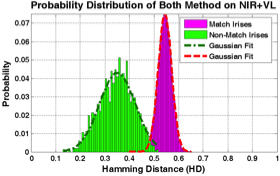}
\label{pdfShapeAhmad}}}
\caption{(a)-(d) Classification results (FAR) for different situations of train \& test data selection. The black dot, blue dash-dot, red dashed, green solid, cyan solid-square, brown solid-circle and pink solid-diamond plots, respectively, determine Poursaberi's Method on NIR, Poursaberi's Method on VL, our proposed method on NIR, our proposed method on VL, Poursaberi's Method on the fusion of NIR+VL, our proposed method on the fusion of NIR+VL and fusion of our method plus Poursaberi's method on NIR+VL datasets; (e) Probability distribution of match and non-match class for Pousaberi's method NIR+VL; (f) Probability distribution for match and non-match class for proposed method NIR+VL (g) Probability distribution for match and non-match class for Pousaberi's and proposed method fusion}
\label{farUIRIS}
\end{figure}

At first step, we used our proposed method on UTIRIS to extract $\textbf{ShapeCode}^{\textbf{Iris}}$ from each eye in both sessions. The number of scenarios for train and test is similar to UBIRIS, Table \ref{UbirisCasia}. TThe results of classifications are shown in Figure \ref{farUIRIS} by red dashed and green solid plots for NIR and VL sessions, respectively. As it is shown, the accuracy results per false acceptance rate (FAR) for VL session suppress the results for NIR session, where the reason goes back to the elimination of pigment melanin in NIR session.

In the next step, we applied a second feature extraction method on UTIRIS to compare the results with our own algorithm. We used Poursaberi's method~\cite{Poursaberi:2007}, where they used Daubechies2 wavelet to extract features from enhanced iris image. The results are shown in Figure \ref{farUIRIS} by black dot and blue dashed-dot plots, where the accuracy on VL session suppresses the NIR session's results due to high content of information on VL session (as discussed above). Although the accuracy results of one false acceptance rate for Poursaberi's method remain higher than our proposed algorithm, the accuracy results of the proposed shape analysis suppress Poursaberi's outcomes at the beginnings of the false acceptance rate, e.g. three false accepted rates (3 FAR). The second advantage of our proposed method is faster convergence with the increase in false acceptance rate.

An important goal of working on a database like UTIRIS is to analyze the correlation of information contained in VL and NIR sessions. In fact, the following argument can be studied on several horizons. One method is to concatenate the feature codes of VL with NIR:
\begin{equation}
\textbf{F}^{\textbf{Concatinated}}=\left[\begin{array}{cc}
\textbf{F}^{\textbf{VL}} & \textbf{F}^{\textbf{NIR}}
\end{array}
\right]
\label{Augm}
\end{equation}
The rationale behind the concatenation is as follows: all features analyzed in VL and NIR have the same characteristics where it comes from the same feature extraction method (Shape analysis and Daubechies2 Wavelet) for both sessions. Our concern in this paper was on Eumelain spectroscopy and its effects on NIR and VL imaging, while we introduce new approach for extracting such related information. In this case we studied complementary information contain in NIR and VL for better results. However, one can perform better fusion analysis for better accuracies, e.g. ~\cite{Yager:2004}.

Recently, Kumar ~\cite{kumar2008} presented a comparative study of the performance from the iris identification using log-Gabor, Haar wavelet, DCT and FFT based features. There results provide higher performance from the Haar wavelet and log Gabor filter based phase encoding. They have implemented their proposed method on CASIA v.1, where the combination of both encoders is most promising, in terms of performance and the computational complexity.

Here, combined features extracted by the shape analysis technique from the NIR and VL sessions. The results of classifications are shown in figure \ref{farUIRIS} with the brown solid-circle. The results of fusion of Poursaberi's method on both sessions are shown by the cyan solid-square, where it can be easily determined that our proposed method with the fusion of data suppresses Poursaberi's method. Besides, the results of both methods are highly improved.

As the last experiment, we fused both methods, adding both dataset on VL and NIR, where the results for classifications are extremely improved. The best answer is obtained from the fourth scenario (4-train and 1-test) by reaching an accuracy of 99.27\% and it completes the classification accuracy to 100\% by accepting more than two false rates (2 FAR), see figure \ref{farUIRIS} the pink solid-diamond plots. We gathered all results for one false acceptance rate (1 FAR) on Table \ref{1farUTIRIS}.

\begin{table}
\renewcommand{\arraystretch}{1}
\caption{1-FAR Classification Results for Different Scenarios Defined on UTIRIS}
\centering
\begin{tabular}{c|c|c|c|c|c|c|c}
 \hline
 \hline
 Scenario & \multicolumn{3}{c}{Shape Analysis Method} & \multicolumn{3}{c}{Wavelet Daubechies2} & Concatination of Both \\
 \hline
 & NIR & VL & NIR+VL & NIR & VL & NIR+VL & NIR+VL \\
 \hline
 1 & 50.33\% & 59.15\% & 77.12\% & 57.19\% & 60.46\% & 73.20\% & 89.22\% \\
 \hline
 2 & 67.62\% & 70.26\% & 87.89\% & 71.59\% & 75.33\% & 85.90\% & 96.26\% \\
 \hline
 3 & 70.95\% & 78.72\% & 90.20\% & 77.36\% & 82.09\% & 90.20\% & 95.95\% \\
 \hline
 4 & 76.71\% & 84.06\% & 94.93\% & 89.13\% & 86.23\% & 96.38\% & 99.28\% \\
 \hline
 \hline
\end{tabular}
\label{1farUTIRIS}
\end{table}

The results of classification for the concatenated case are highly improved. The features contained in both sessions present different patterns and textures. This consequence was almost predictable, where we noted in previous sections that NIR imaging eliminates most pigment melanin due to lack of emission by the following macromolecules under NIR firing. We realized that the textures previewed in NIR imaging are mostly related to soft tissues of the iris, called sphincter and radial muscles~\cite{Guyton:2000}.

Probability distributions of fusion cases for Poursaberi's method, the proposed method and the augmentation of both are shown in Figure \ref{pdfAhmad}-\ref{pdfShapeAhmad}, where the distribution of match-class and non-match-class leads to good separability in data fusion. Figure \ref{utiris4} indicates the results of fusing data, where the correct classification rate is highly improved. We get 99.28\% accuracy in the first step of false acceptance and reach 100\% accuracy in the second step. This shows that data fusion can be the key issue for the scalability of Iris recognition methods to larger databases. This analysis provides some leads towards classification problems in large size databases.

\section{Conclusion}
Our proposed method is highly resistant to noise in the VL images. The proposed algorithm encodes the pattern of pigment melanin in the VL image independent of textures in the NIR image. It also extracts invariant features from the VL and the NIR images whose fusion leads to higher classification accuracy. However, this method uses a large strip-code in the order of thousands of bits contrary to the previous methods. For example,  Daugman uses 2,048-bits codes while our code is in the order of 10 thousands (e.g., 19,200 bits). Three features were proposed: RVF; SF; and TAF. Some other facts, e.g. number of shapes, number of samples (N) can be variable due to complexity of each defined binary template. As a conclusion, VL imaging should be considered to trusted-zones of iris biometrics where the patterns of pigment melanins are highly meaningful and can produce a valuable encoded data for classification and complementary features to NIR images.

\section*{Acknowledgment}
The authors would like to thank Soft Computing and Image Analysis Group from University of Beira Interior-Portugal for use of the UBIRIS Iris Image Database. Portions of the research in this paper use the CASIA-IrisV1 collected by the Chinese Academy of Sciences' Institute of Automation (CASIA). They would like to specially thank Mr. Ahmad Poursaberi for his helpful discussions and providing some results. The first author also would like to thank Ms. Mahta Karimpoor and Mr. John Varden for their helpful comments. This work was supported by the Control and Intelligent Processing Centre of Excellence (CIPCE) at the University of Tehran and Iran Telecommunication Research Center (ITRC).


\begin{biography}[{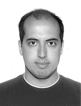}]{Mahdi S. Hosseini}
(S'08) was born in Tabriz, Iran, in 1983. He received the BSc degree in electrical engineering from the University of Tabriz in Iran in 2004. He then entered the University of Tehran and received the MSc degree in electrical engineering in 2007. He is currently pursuing his PhD in ECE department of University of Waterloo since Feb 2008 under supervision of Prof. Oleg Michailovich. His main interest includes machine vision, pattern recognition particularly applied in iris biometrics, sparse signal recovery, compressed sensing, 2D-phase unwrapping and	multiresolution analysis.
\end{biography}

\begin{biography}[{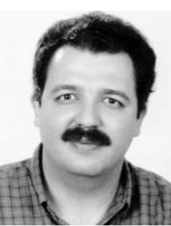}]{Babak N. Araabi} 
(S'98-M'01) was born in 1969. He received the B.S. degree from the Sharif University of Technology, Tehran, Iran, in 1992, the M.S. degree from University of Tehran, Tehran, in 1996, and the Ph.D. degree from Texas A\&M University, College Station, in 2001, all in electrical engineering. In January 2002, he joined the Department of Electrical and Computer Engineering, University of Tehran, where he is now an Associate Professor and Head of the Control Systems Group. He is also a Research Scientist with the School of Cognitive Sciences, Institute for Studies in Theoretical Physics and Mathematics, Tehran. He is the author or coauthor of more than 100 international journals and conference papers in his research areas, which include machine learning, pattern recognition, neuro-fuzzy modeling, prediction, predictive control, and system modeling and identification.
\end{biography}

\begin{biography}[{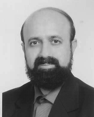}]{Hamid Soltanian-Zadeh} 
(S'90-M'92-SM'00) was born in Yazd, Iran in 1960.  He received BS and MS degrees in electrical engineering: electronics from the University of Tehran, Tehran, Iran in 1986 and MSE and PhD degrees in electrical engineering: systems and bioelectrical sciences from the University of Michigan, Ann Arbor, Michigan, USA, in 1990 and 1992, respectively.  Since 1988, he has been with the Department of Radiology, Henry Ford Hospital, Detroit, Michigan, USA, where he is currently a Senior Scientist and director of Medical Image Analysis Group. Since 1994, he has been with the Department of Electrical and Computer Engineering, the University of Tehran, Tehran, Iran, where he is currently a full Professor and director of Control and Intelligent Processing Center of Excellence (CIPCE). Prof. Soltanian-Zadeh has active research collaboration with the University of Michigan, Ann Arbor, and Wayne State University, Detroit, MI, USA, and the Institute for studies in theoretical Physics and Mathematics (IPM), Tehran, Iran. His research interests include medical imaging, signal and image processing and analysis, pattern recognition, and neural networks.  He has co-authored over 450 papers in journals and conference records or as book chapters in these areas. Several of his presentations received honorable mention awards at the SPIE and IEEE conferences. In addition to the IEEE, he is a member of the SPIE, SMI, and ISBME and has served on the scientific committees of several national and international conferences and review boards of about 40 scientific journals. Prof. Soltanian-Zadeh is a member of the Iranian Academy of Sciences and has served on the study sections of the National Institutes of Health (NIH), National Science Foundation (NSF), American Institute of Biological Sciences (AIBS), and international funding agencies.
\end{biography}

\end{document}